\documentclass{llncs}
\usepackage[T1]{fontenc}
\usepackage{url}
\usepackage{amsmath}
\usepackage{graphicx}
\usepackage{array}
\begin{document}

\title{Bank Card Usage Prediction Exploiting Geolocation Information}
%\titlerunning{Two-Stage Transfer Surrogate Model}
\author{Martin Wistuba \and Nghia Duong-Trung \and Nicolas Schilling \and Lars Schmidt-Thieme}
\authorrunning{Martin Wistuba et al.}
\institute{Information Systems and Machine Learning Lab \\
University of Hildesheim\\
Universit\"atsplatz 1, 31141 Hildesheim, Germany  \\
\email{\{wistuba,duongn,schilling,schmidt-thieme\}@ismll.uni-hildesheim.de}
}

\maketitle              % typeset the title of the contribution

\begin{abstract}
We describe the solution of team \textit{ISMLL} for the ECML-PKDD 2016 Discovery Challenge
on Bank Card Usage for both tasks. Our solution is based on three
pillars. Gradient boosted decision trees as a strong regression and
classification model, an intensive search for good hyperparameter
configurations and strong features that exploit geolocation information.
This approach achieved the best performance on the public leaderboard
for the first task and a decent fourth position for the second task.
\end{abstract}

\section{Challenge Description}

The goal of one of this year's ECML-PKDD Discovery Challenges was
to predict the behaviour of customers of the Hungarian bank \textit{otpbank}. The challenge
was divided into two tasks. The first task was to predict for every
bank branch the number of visits for a set of customers, the second
task was to predict, whether a customer will apply for a credit card in
the next six months. For these tasks, anonymized customer information
(e.g. age, location, income, gender) and bank activities (e.g. what has been bought, where and when) were provided.
A labeled data set for 2014 was made available which can be used
for supervised machine learning to predict the targets for a disjoint
set of customers for 2015. The evaluation measure for Task 2 is the
area under the ROC curve (AUC), a very common measure for imbalanced classification
problems. The evaluation measure for Task 1 is a little bit more exotic.
It is the average of cosine@1 and cosine@5 for every customer $c$
where 
\begin{equation}
\text{cosine@k}:=\frac{\sum_{i=1}^{k}y_{c,i}\hat{y}_{c,i}}{\sqrt{\sum_{i=1}^{b}y_{c,i}^{2}}\sqrt{\sum_{i=1}^{k}\hat{y}_{c,i}^{2}}}
\end{equation}
with $y_{c,i}$ being the number of times the customer $c$ has visited
bank branch $i$ and $\hat{y}_{c,i}$ the
prediction, respectively.  There are $b$ different branches in total.
For more information we refer to the challenge
website \cite{ChallengeWebsite}.

\section{Problem Identification}

For the first task, we assumed that there is no relation between the
number of visits of a customer among branches. This
enabled us to tackle $b$ different regression tasks for each of the
$b$ branches. Independently, we trained a regression model for each
branch that predicts for a customer how often she will visit that
branch based on past information for that branch. This is a classical
example for count data and hence, we tackled this task as a Poisson
regression problem. For Task 1 we had to select five bank branches for which we wanted
to make predictions. We simply chose the five with highest predicted
number of visits which is the best way to achieve a good score in
case the predictor performs reasonable.

We considered Task 2 to be a classification task. We minimized
the logistic loss and considered the class imbalance
by choosing an appropriate class weight.

For both tasks, we used gradient boosted decision trees \cite{Chen2016}
as the prediction model.

\section{Data Preprocessing}

For the feature and hyperparameter selection we had to split the labeled
data set into a training data set $D_{\text{train}}$ and a validation
data set $D_{\text{valid}}$ such that the performance on $D_{\text{valid}}$
will reflect the performance on the hidden test data. The task was
to infer from some customers and their activities in 2014 
the behaviour of a disjoint set of customers in 2015. Only basic customer
information as well as the customer's activities of the first half
of 2015 (excluding branch visits) was given for the test customers.
Thus, we decided to split the given labeled data set by customers,
selecting 80\% for $D_{\text{train}}$ and the remaining 20\% for
$D_{\text{valid}}$ uniformly at random. Only the first six months
of activities of the validation customers (excluding branch visits)
was provided for validation purposes. The only problem here is that
we are actually predicting from data from 2014 for customers in 2014
but there was no way to overcome this problem.

Very basic information of the customers was available including age,
location, income and gender. While gender is by nature binary, the
other features were already binned into three categories. We employed
this information as features after transforming them via one-hot encoding.
Furthermore, the internal classification of a bank whether the customer
is considered as wealthy or not was given for each month.
We distinguished
customers of following five categories: customers that have been classified
as 1) wealthy in all observed months, 2) not wealthy in all observed
months, 3) first wealthy and then changed to not wealthy, 4) first
not wealthy and then changed to wealthy, 5) those who changed their
classification more than once. Applying one-hot encoding, we added
this information as features.

Finally, the information in what month the customer possesses a credit
card of the bank was provided. Analogously to the five categories
of the wealthy classification, we created categories for the credit
card time-series information.

\begin{figure}
\includegraphics[width=1\textwidth]{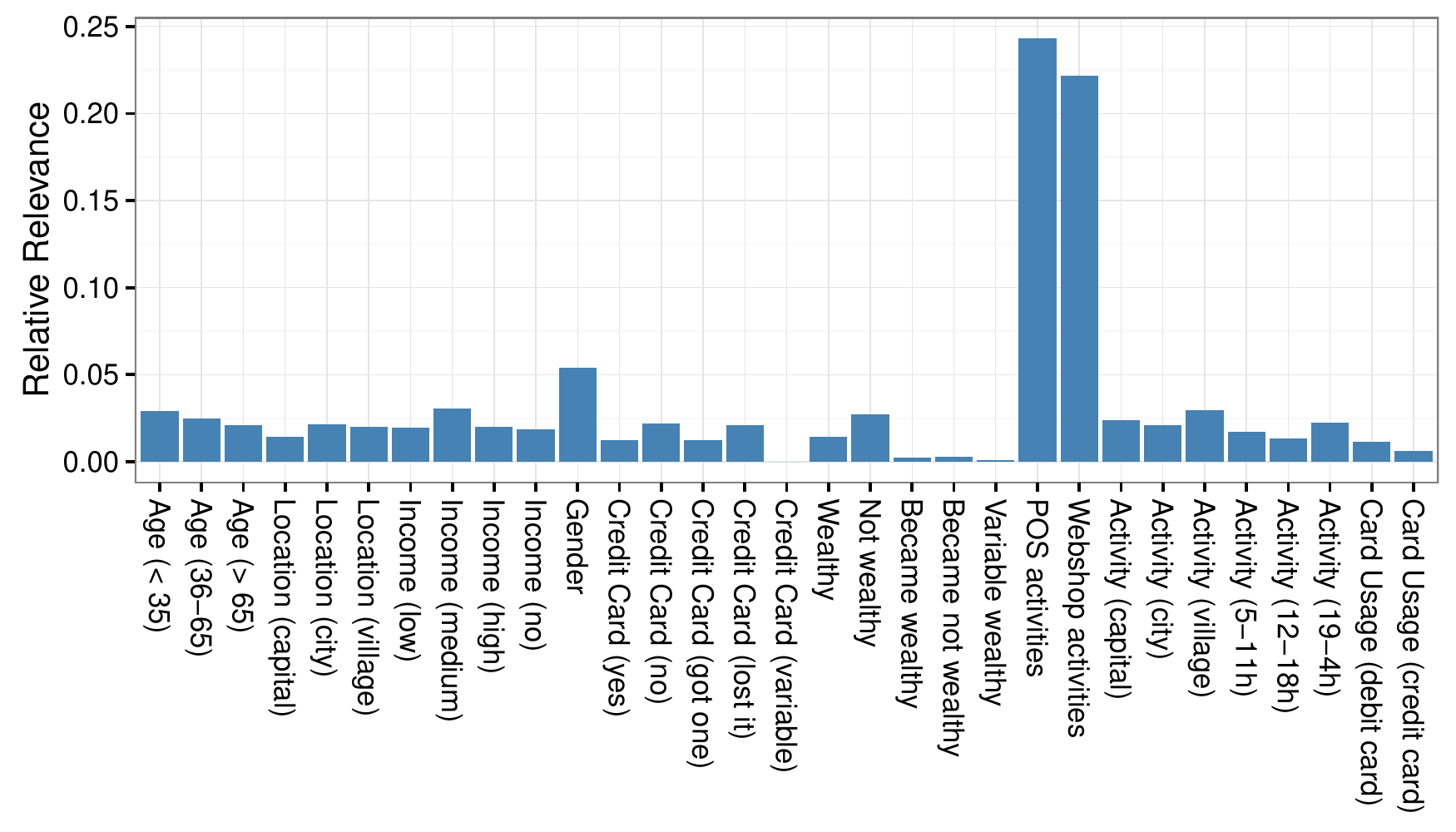}
\protect\caption{This plot visualizes the relative number of times a feature was chosen to build a tree in Task 2. 
The activity features are used in almost every fourth tree.\label{fig:feature_importance2}}
\end{figure}

Besides using basic customer features, we wanted to use the information
of the customer's activities. While we found many features that improved
the performance for Task 2 on our internal data split, we saw for
many features no improvement on the public leaderboard. Thus, the
only feature we used is the number of activities per channel committed
by the customer. Figure \ref{fig:feature_importance2} shows that it
is one of the most predictive features.

\begin{figure}
\includegraphics[width=1\textwidth]{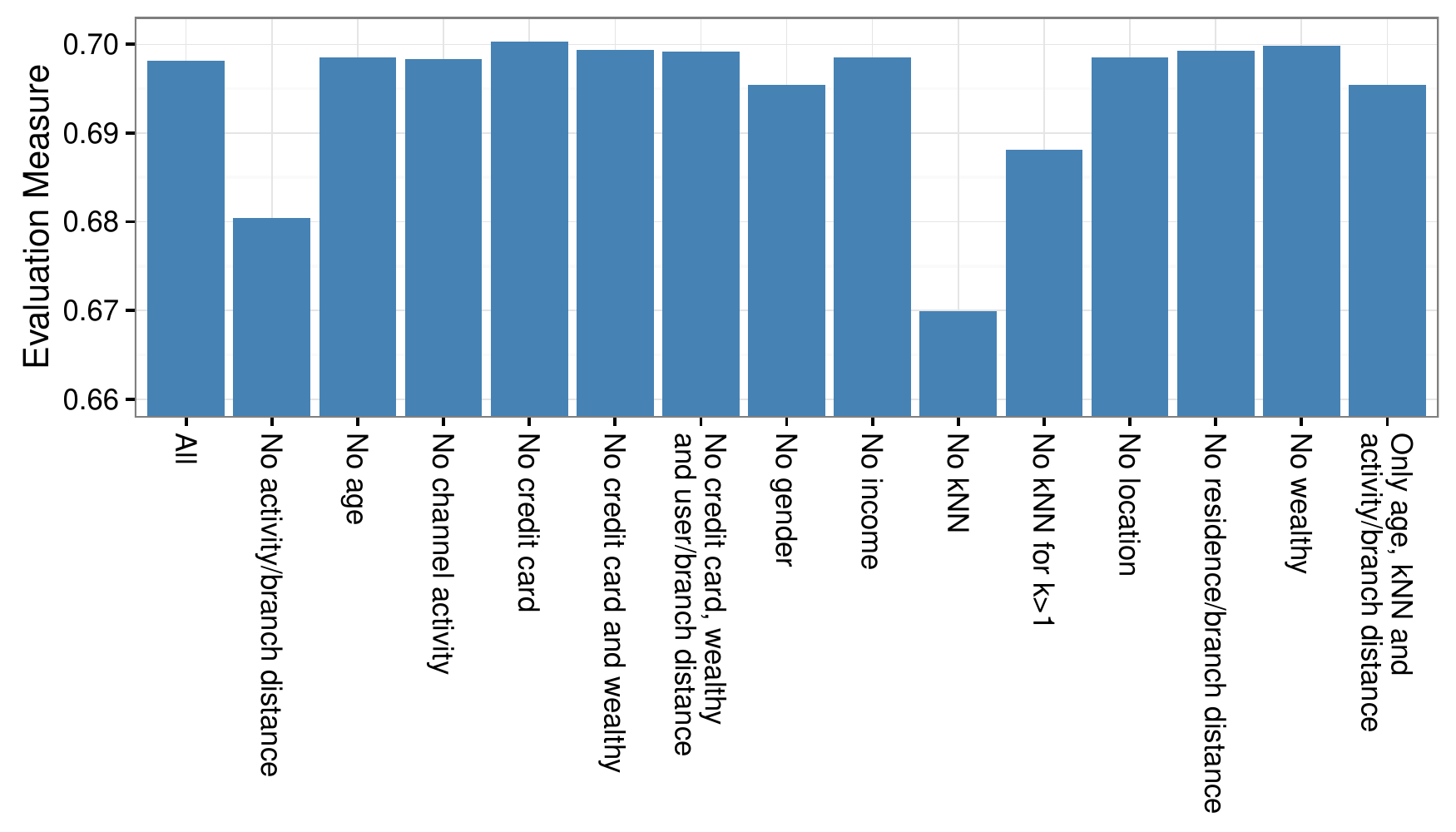}
\protect\caption{Intermediate feature backward selection results for Task 1. Location-aware
features provide huge improvements.\label{fig:feature_selection}}
\end{figure}

\begin{figure}
\includegraphics[width=1\textwidth]{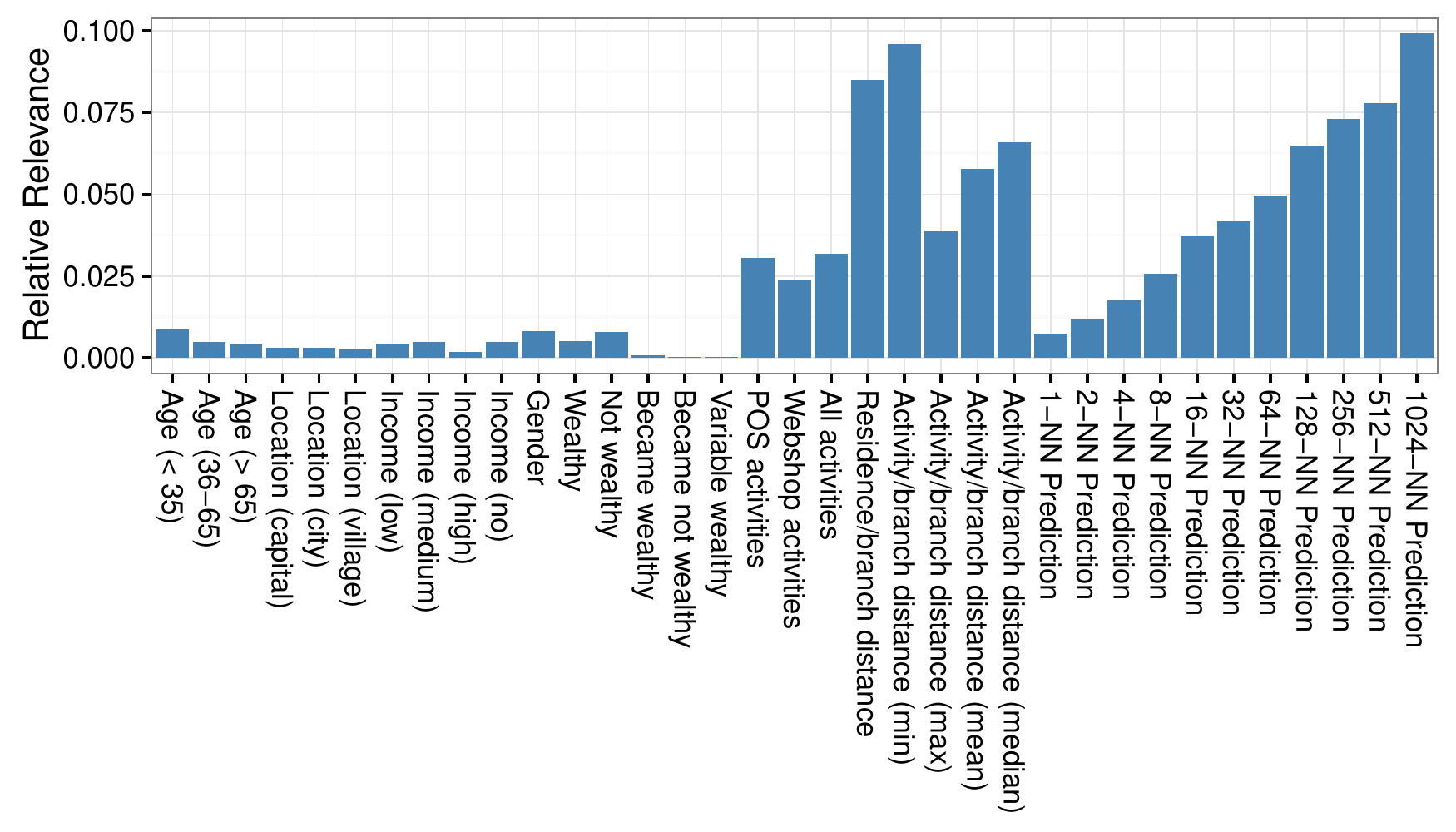}
\protect\caption{This plot visualizes the relative relevance of all features used in Task 1. The
higher the score, the more often the feature was used for building a tree. Location-aware features
prove to be highly predictive.\label{fig:feature_importance1}}
\end{figure}

For Task 2 we considered location information about activities, bank
branches and customers to be irrelevant and only used aforementioned
features. However, for Task 1 this information was one of the most
impactful information. One feature we used was the distance
between the residence of the customer and a bank branch which is a
quite obvious choice. Digging into the data, we saw that there were
many customers using bank branches very far away from their residence.
We tried to cover this by also adding the maximum, minimum, mean and
median distance between a bank branch and the customer's activities.
Finally, we added k-nearest-neighbors predictions for $k=2^{0},2^{1},\ldots,2^{10}$
using the Euclidean distance between the residence of customers as
the distance function. These features follow the simple assumption
that customers that live nearby visit the same bank branches. Figure
\ref{fig:feature_selection} provides insight into our intermediate
feature selection experiments for Task 1 and clearly shows the importance
of the location-aware features. Based on this experiment, we used
all features but the credit card information for Task 1.
Figure \ref{fig:feature_importance1} shows the relative frequency
of a specific feature being taken as a splitting variable.
Again, this shows the importance of location-aware features for Task 1.

\section{Hyperparameter Tuning and Ensembling}

For both tasks we tuned the hyperparameters by considering the choice
of hyperparameters $\boldsymbol{\lambda}$ as a black-box optimization
problem
\begin{equation}
\underset{\boldsymbol{\lambda}}{\arg\min\ }\mathcal{L}\left(\hat{y}_{\boldsymbol{\lambda}}\left(D_{\text{valid}}\right),y_{\text{valid}}\right)
\end{equation}
where $\hat{y}_{\boldsymbol{\lambda}}$ is the model that was trained
on the training partition of the data $D_{\text{train}}$ using hyperparameter
configuration $\boldsymbol{\lambda}$ and $\hat{y}_{\boldsymbol{\lambda}}\left(D_{\text{valid}}\right)$
the corresponding predictions for the validation partition $D_{\text{valid}}$.
Then, the problem of hyperparameter tuning is to find a hyperparameter
configuration $\boldsymbol{\lambda}$ such that a loss function $\mathcal{L}$
given the predictions and the groundtruth is minimized.

\begin{figure}
\includegraphics[width=1\textwidth]{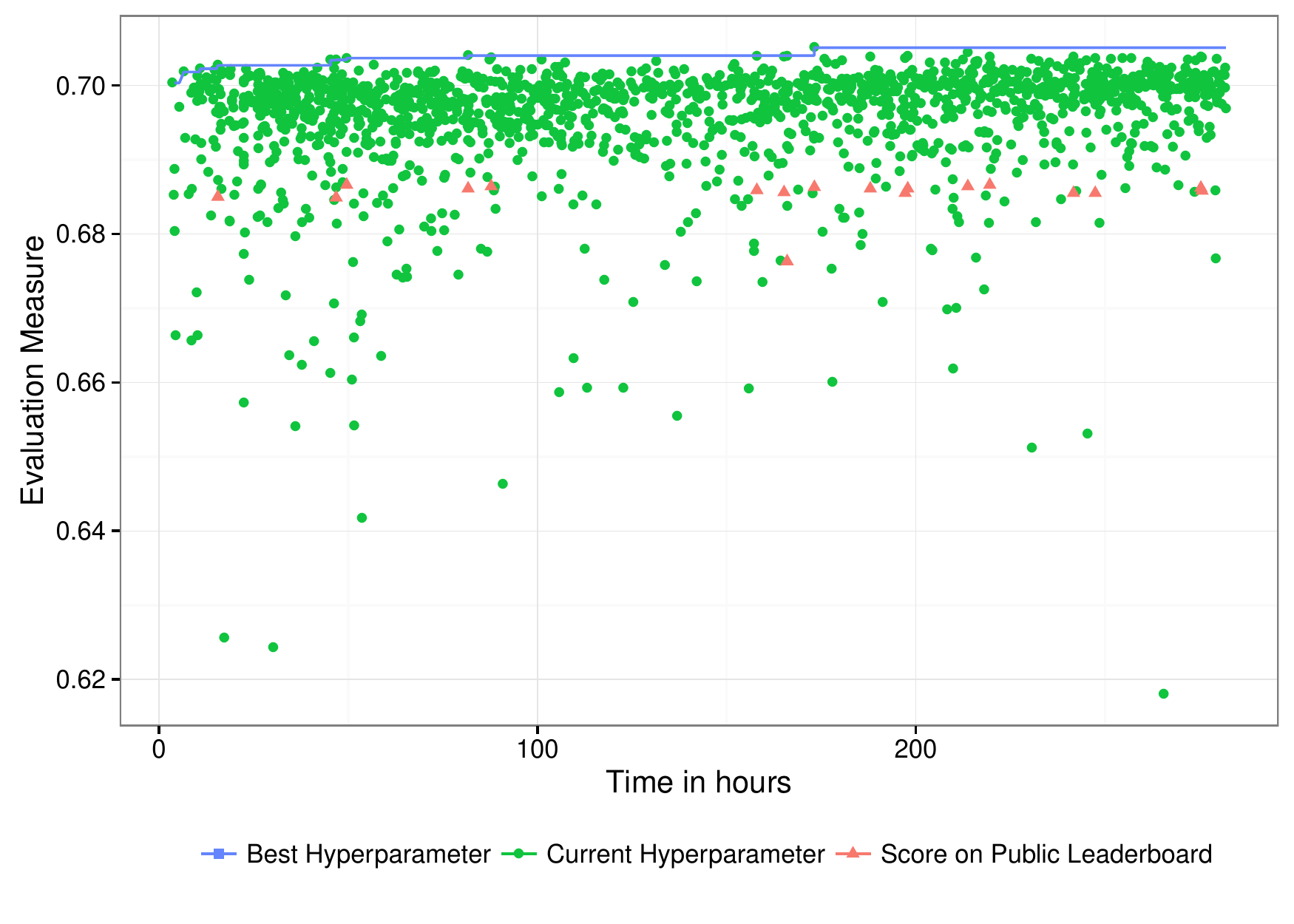}
\protect\caption{Searching for a good hyperparameter configuration with 100 cores in
parallel. The public leaderboard score is shown for some of the best
hyperparameter configurations on our validation set.\label{fig:hp-search}}
\end{figure}

We tackled this black-box optimization problem using Sequential Model-based
Optimization (SMBO) \cite{Snoek2012}. Figure \ref{fig:hp-search}
presents the progress of the optimization process that was conducted
in parallel on 100 cores for our own train/validation split as well
as results on the public leaderboard for Task 1.

For Task 2, we tried diverse ways of ensembling using different
base models but did not achieve any improvement. In the end, we averaged
the predictions of 100 models for the best hyperparameter configuration
using different seeds.

One of our best single models for Task 1 achieved a cosine@1 score of $0.676$
and a cosine@5 score of $0.728$ leading to an overall score of $0.702$
on our validation split. The performance on the test set is with $0.686$
much smaller. A possible reason might be temporal effects because the predictions
for the test customers are for 2015 but we learn on data from 2014.

\begin{table}
\protect\caption{The performance of the top five teams for both tasks on the public
leaderboard.}
\centering{}%
\begin{tabular}{p{3.2cm}>{\raggedleft}p{1.5cm}p{1cm}p{3.2cm}>{\raggedleft}p{1.5cm}}
\cline{1-2} \cline{4-5} \noalign{\smallskip}
\multicolumn{2}{c}{Task 1} &  & \multicolumn{2}{c}{Task 2}\tabularnewline
Team & Score &  & Team & Score\tabularnewline
\noalign{\smallskip}
\cline{1-2} \cline{4-5} 
\noalign{\smallskip}
\textbf{1. ISMLL} & \textbf{0.68659} &  & 1. achm & 0.71862\tabularnewline
2. Ya & 0.68512 &  & 2. Cosine Vinny & 0.71730\tabularnewline
3. Cosine Vinny & 0.67436 &  & 3. Degrees of Freedom & 0.71589\tabularnewline
4. Outliers & 0.65607 &  & \textbf{4. ISMLL} & \textbf{0.71523}\tabularnewline
5. seed71 & 0.65287 &  & 5. TwoBM & 0.71479\tabularnewline
\cline{1-2} \cline{4-5} 
\end{tabular}
\end{table}

\section{Conclusions}

Our solution is based on strong ensemble methods, smart feature engineering
and an intense search for optimal hyperparameter configurations. For
Task 1, this paid of leading to the best solution regarding the public
leaderboard as well as a decent result for Task 2.

\subsubsection*{Acknowledgments.}

The authors gratefully acknowledge the co-funding of their work 
by the German Research Foundation (DFG) under grant SCHM 2583/6-1.

\bibliographystyle{splncs03}
\bibliography{references} 
\end{document}